%% file: main.tex
\documentclass{article}

\PassOptionsToPackage{numbers, compress}{natbib}
    \usepackage[preprint]{neurips_2025}

\usepackage[utf8]{inputenc} % allow utf-8 input
\usepackage[T1]{fontenc}    % use 8-bit T1 fonts
\usepackage{hyperref}       % hyperlinks
\usepackage{url}            % simple URL typesetting
\usepackage{booktabs}       % professional-quality tables
\usepackage{amsfonts}       % blackboard math symbols
\usepackage{nicefrac}       % compact symbols for 1/2, etc.
\usepackage{microtype}      % microtypography
\usepackage{xcolor}         % colors
\usepackage{enumitem}
\usepackage{graphicx}
\usepackage{xspace}
\usepackage{amsmath}
\usepackage{cleveref}
\usepackage{mathrsfs}
\usepackage{mathdesign}
\usepackage{hyperref}
\usepackage{wrapfig}

\crefname{figure}{Fig.}{Figs}
\crefname{table}{Tab.}{Tabs}

\title{ViewPoint: Panoramic Video Generation with Pretrained Diffusion Models}

\author{
  Zixun Fang$^{1,2}$\thanks{Intern at Tongyi Lab.} \quad Kai Zhu$^{1,2}$ \quad Zhiheng Liu$^{3}$ \quad Yu Liu$^{2}$ \quad Wei Zhai$^{1\dag}$ \\  \textbf{Yang Cao}$^{1}$ \quad \textbf{Zheng-Jun Zha}$^{1}$\\
\\
  $^1$ USTC \quad
  $^2$ TongYi Lab \quad
  $^3$ HKU 
}

\def\ie{{\it{i.e.,~}}}
\def\eg{{\it{e.g.,~}}}

\def\method{ViewPoint}

\begin{document}

\maketitle
\begin{figure}[!h]
  \centering
    \vspace{-1.0cm}
  \includegraphics[width=1\linewidth]{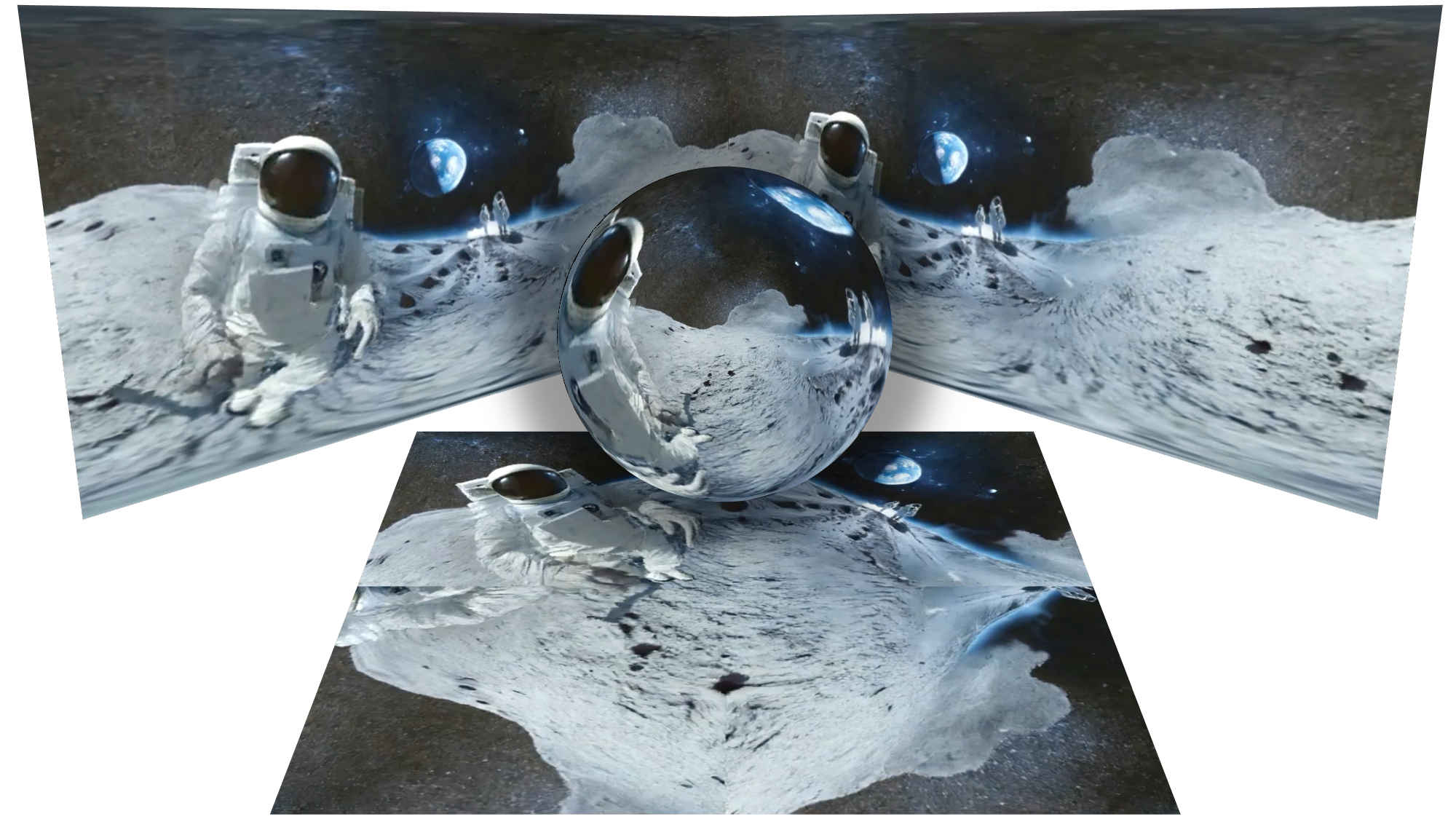}
  \caption{\textbf{The generated result.} The image at the bottom displays the ViewPoint map generated by our model, with the background image showing the concatenated equirectangular map derived from the ViewPoint map. The generated panoramic image exhibits excellent spatial consistency, as the equirectangular map can be seamlessly stitched together. %For video results, please refer to the anonymous link: \href{https://anonymouser00.github.io/}%{https://anonymouser00.github.io/}
  Project page:  \href{https://becauseimbatman0.github.io/ViewPoint}{\textcolor{red}{ViewPoint}}.
  }
  \label{fig:teaser}
  %\vspace{-0.5cm}
\end{figure}

\input{secs/0_abs}
\input{secs/1_intro}
\input{secs/2_related}

\input{secs/3_method}
\input{secs/4_exp}
\input{secs/5_conclusion}
\input{secs/6_ref}

\end{document}

%% file: secs/0_abs.tex
\begin{abstract}
Panoramic video generation aims to synthesize 360-degree immersive videos, holding significant importance in the fields of VR, world models, and spatial intelligence. 
Existing works fail to synthesize high-quality panoramic videos due to the inherent modality gap between panoramic data and perspective data, which constitutes the majority of the training data for modern diffusion models.
In this paper, we propose a novel framework utilizing pretrained perspective video models for generating panoramic videos.
Specifically, we design a novel panorama representation named ViewPoint map, which possesses global spatial continuity and fine-grained visual details simultaneously.
With our proposed Pano-Perspective attention mechanism, the model benefits from pretrained perspective priors and captures the panoramic spatial correlations of the ViewPoint map effectively.
Extensive experiments demonstrate that our method can synthesize highly dynamic and spatially consistent panoramic videos, achieving state-of-the-art performance and surpassing previous methods.

% Kuai Ma Jia Bian Kuai Ma Jia Bian Kuai Ma Jia Bian Kuai Ma Jia Bian Kuai Ma Jia Bian Kuai Ma Jia Bian Kuai Ma Jia Bian Kuai Ma Jia Bian Kuai Ma Jia Bian Kuai Ma Jia Bian Kuai Ma Jia Bian Kuai Ma Jia Bian Kuai Ma Jia Bian Kuai Ma Jia Bian Kuai Ma Jia Bian Kuai Ma Jia Bian Kuai Ma Jia Bian Kuai Ma Jia Bian Kuai Ma Jia Bian Kuai Ma Jia Bian Kuai Ma Jia Bian Kuai Ma Jia Bian Kuai Ma Jia Bian Kuai Ma Jia Bian Kuai Ma Jia Bian Kuai Ma Jia Bian Kuai Ma Jia Bian Kuai Ma Jia Bian Kuai Ma Jia Bian Kuai Ma Jia Bian Kuai Ma Jia Bian Kuai Ma Jia Bian Kuai Ma Jia Bian Kuai Ma Jia Bian Kuai Ma Jia Bian Kuai Ma Jia Bian Kuai Ma Jia Bian Kuai Ma Jia Bian Kuai Ma Jia Bian Kuai Ma Jia Bian Kuai Ma Jia Bian Kuai Ma Jia Bian Kuai Ma Jia Bian Kuai Ma Jia Bian Kuai Ma Jia Bian Kuai Ma Jia Bian Kuai Ma Jia Bian Kuai Ma Jia Bian  
\end{abstract}

%% file: secs/1_intro.tex
\section{Introduction}
Imagine you're traveling—would you seize the chance to capture the breathtaking landscapes that unfold before you? When you return home, do you long to relive and immerse yourself in those vivid experiences once more?
Recently, omnidirectional vision has garnered increasing attention as it unlocks immersive AR/VR, virtual travel, and telepresence experiences.
However, recording 360-degree videos requires expensive professional devices, \ie 360 cameras, making the creation of panoramic content challenging, and most consumers capture only narrow field-of-view (FoV) clips on portable monocular cameras, \eg smartphones.
Enabling these perspective recordings to become full panoramas would democratize spherical media, letting anyone relive or share memories in true 360-degree form.

Achieving this conversion is non-trivial due to fundamental representation gaps between perspective and panorama domains. A straightforward approach is to adopt the widely-used Equirectangular Projection (ERP) for panoramas, which maps the spherical view onto a rectangular image.
Unfortunately, this format introduces severe distortions, especially near the poles, stretching and squashing content unnaturally. More critically, equirectangular images lie outside the distribution of typical training data for modern generative models. Diffusion models~\cite{ho2020denoising, rombach2022high,song2021denoising} and VAEs~\cite{kingma2013auto} are predominantly trained on perspective imagery, so without significant adaptation they struggle to produce high-quality results in the warped ERP space.
On the other hand, one could represent the panorama as multiple perspective projections. A common choice is the Cubemap Projection(CP) format, which unfolds the sphere into six faces (each a 90° FOV perspective view). The CP representation avoids polar distortions and yields locally planar patches well-aligned with the priors of convolutional and diffusion networks.
However, a naive cubemap suffers from spatial discontinuities at the borders of the faces. The six faces are disjoint on a 2D grid, making it difficult for a neural model to capture cross-face consistency. 
In short, existing representations force a trade-off between continuity and distortion: ERP offers end-to-end continuity but distorts the content, whereas CP preserves local fidelity but fragments the panoramic space. Neither is ideal for generative video modeling, especially when temporal consistency is also required.

% our method
In this paper, we address the above-mentioned issues by introducing ViewPoint, a novel 360° video representation and generation framework that bridges the strengths of ERP and CP. 
At the core of ViewPoint is a spatially-aware pseudo-stitching scheme that reprojects and rearranges the scene into an overlapping set of perspective views with greatly improved continuity.
Specifically, we first convert the panorama into six cube faces—each a distortion-free perspective view—and then merge them into a small number of overlapping ``pseudo-perspective" panels. 
Because these panels share content along their boundaries, they eliminate the hard seams of a standard cubemap while remaining fully compatible with existing 2D diffusion models.
On this representation, the proposed Pano-Perspective attention delivers two key benefits: (1) Global coherence: Pano-attention blocks span the entire stitched map (and time), ensuring that opposite directions align and the scene stays consistent. (2) Local fidelity: Perspective-attention blocks focus on each panel’s neighborhood, preserving fine texture, color, and motion details.
With this attention mechanism, we take full advantage of the generative capabilities of diffusion models and adapt them to the task of panoramic video generation, where high-quality data is scarce.

In summary, our contributions are as follows: 
\begin{itemize}[leftmargin=*]
\item We introduce a novel representation for 360-degree content aimed at improving spatial continuity and reducing distortion while leveraging the power of diffusion models to generate 360-degree videos through the proposed format.
\item We design a Pano-Perspective attention mechanism, enabling the model to simultaneously maintain global spatial continuity across the entire panorama and significantly enhance the preservation of fine-grained details and motion.
\item Extensive experiments demonstrate that our method can generate high-quality 360-degree videos and achieve state-of-the-art performance, outperforming previous approaches.
\end{itemize}

%% file: secs/2_related.tex
\section{Related Works}
\noindent\textbf{Panorama Representations.} Panoramic images are often projected from spherical space to the 2D plane for processing and storage~\cite{ai2025survey}. 
\textbf{Equirectangular Projection (ERP)} is the most popular format, which uniformly maps pixels from a spherical surface to a planar rectangle.
Despite its simplicity and spatial continuity, ERP inevitably brings disadvantages such as geometric distortion at the poles.
\textbf{Cubemap Projection (CP)}, on the other hand, projects panoramas to six cube faces with the FoV of $90^\circ\times90^\circ$ to alleviate geometric distortion. 
However, under this representation, only one cube face is explicitly spatially continuous with its four adjacent faces on the 2D plane.

\noindent\textbf{Panorama Generation.} Panoramic image generation~\cite{kalischek2025cubediff,Tang2023mvdiffusion, wu2023panodiffusion, Ye2024DiffPano, feng2023diffusion360, chen2022text2light, liu2024panofree, panfusion2024, wang2023360, shum2023conditional, wang2024customizing, li2023panogen} aims to synthesize 360-degree immersive images based on user-provided textual or visual clues. 
It requires the generated images to be seamless in spherical space while maintaining rationality in any perspective anchor. 
Recent studies~\cite{kalischek2025cubediff,Tang2023mvdiffusion, wu2023panodiffusion, Ye2024DiffPano, panfusion2024, wang2023360, wang2024customizing, li2023panogen} leverage powerful image diffusion models to generate high-quality panoramic images. 
Benefiting from the great generative capabilities of diffusion models, these works have made significant advancements in omnidirectional image synthesis.
Among them, PanoGen~\cite{li2023panogen} achieves high-quality results in indoor scenarios by introducing a recursive outpainting and stitching mechanism. 
PanFusion~\cite{panfusion2024} proposes a dual-branch pipeline to integrate equirectangular features and perspective features.
CubeDiff~\cite{kalischek2025cubediff} uses an alternative representation, cubemap, to synthesize panoramic images with high fidelity and diversity.
Despite the significant progress in omnidirectional image generation, achieving similar success in the field of video remains a challenging problem, as it requires spatial-temporal consistency across the entire spherical space, as well as costly computational resources.

A few works~\cite{wang2024360dvd, li20244k4dgen, ma2024vidpanos, tan2024imagine360} explore panoramic video generation. 
4K4DGEN~\cite{li20244k4dgen} animates a given high-resolution panoramic image through user intention.
However, the requirement of high-quality panoramic inputs limits the application scenarios, as it struggles to generate omnidirectional content without the given images.
VidPanos~\cite{ma2024vidpanos} extends a given video to a larger FoV by using Temporal Coarse-to-Fine and Spatial Aggregation strategies,  but it does not produce $360^\circ\times180^\circ$ FoV panoramas.
360DVD~\cite{wang2024360dvd} trains a 360-Adapter to exploit the generative capabilities of text-to-video diffusion models, achieving text-driven 360-degree video generation.
Most relevant to our work, Imagine360~\cite{tan2024imagine360} utilizes a dual-branch design, similar to~\cite{panfusion2024}, to synthesize panoramic videos from given perspective inputs. 
However, the equirectangular videos generated by both 360DVD~\cite{wang2024360dvd} and Imagine360~\cite{tan2024imagine360} exhibit severe distortion, especially at the poles, significantly degrading the sense of immersion and realism.
We argue that the spatial distortion and motion drifting caused by ERP make it difficult for models to effectively understand the polar regions with limited training data. 
After all, generating perspective videos with extreme motion is still a challenge, let alone panoramas.

\noindent\textbf{Video Outpainting.} Formally, extending from a given perspective view to a panoramic video is a kind of video outpainting task.
Benefiting from pretrained diffusion models~\cite{ho2020denoising,song2020denoising,rombach2022high,guo2023animatediff}, previous works~\cite{wang2024your,chen2024follow} have made notable progress in perspective video outpainting. 
However, due to the scarcity of 360-degree video data and the modality gap introduced by its unique representation form, panoramic video outpainting remains an open question.
In this paper, we design a novel panorama representation format to fully exploit the generative priors of video diffusion models, aiming to address the absence of high-quality panoramic video outpainting.

%% file: secs/3_method.tex
\section{Method}
\subsection{Preliminary}
Latent diffusion models~\cite{rombach2022high, ho2020denoising, peebles2023scalable} conduct a series of diffusion and denoising processes in latent space. 
Given a clean latent code $x_0$ from the training data, a noisy latent $x_t$ is obtained by adding an random noise $x_0\sim \mathcal{N}(0, I)$ to $x_0$ according to a timestep $t\in [0, 1]$. 
Following the flow matching paradigm~\cite{lipman2023flow,esser2024scaling}, $x_t$ is defined as
\begin{equation}  
x_t = t x_1 + (1 - t) x_0.
\end{equation}
The training objective of the model is to predict the velocity $v_t$, thus, the loss function of the training process can be formulated as
\begin{equation}  
L = \mathbb{E}_{x_0, x_1, c_{txt}, t} \left\lVert u(x_t, c_{txt}, t; \theta) - v_t \right\rVert^2,
\end{equation}
where $c_{txt}$ is the text condition, $\theta$ is the model weights, $u$ is the predicted velocity and $v_t$ is the ground truth velocity.

% \subsection{ERP or CP ?}
% Latent diffusion models encode the input data into latent space through VAEs.
% %
% Therefore, we first examine the reconstruction quality for the two representations, ERP and CP, to reveal which representation is better to choose within the modern diffusion paradigm.
% %
% The details of the experiments are given in Section 4. 
% %
% The experiments show that CP has better reconstruction quality compared to ERP. 
% %
% However, CP lacks an explicit representation of global spatial information. 
% %
% For example, deducing a specific 3D object from given three orthographic views is not an easy task, while the panorama spatial information in CP is implicitly contained within each face.

\begin{figure}[!h]
  \centering
    % \vspace{-1.0cm}
  \includegraphics[width=1\linewidth]{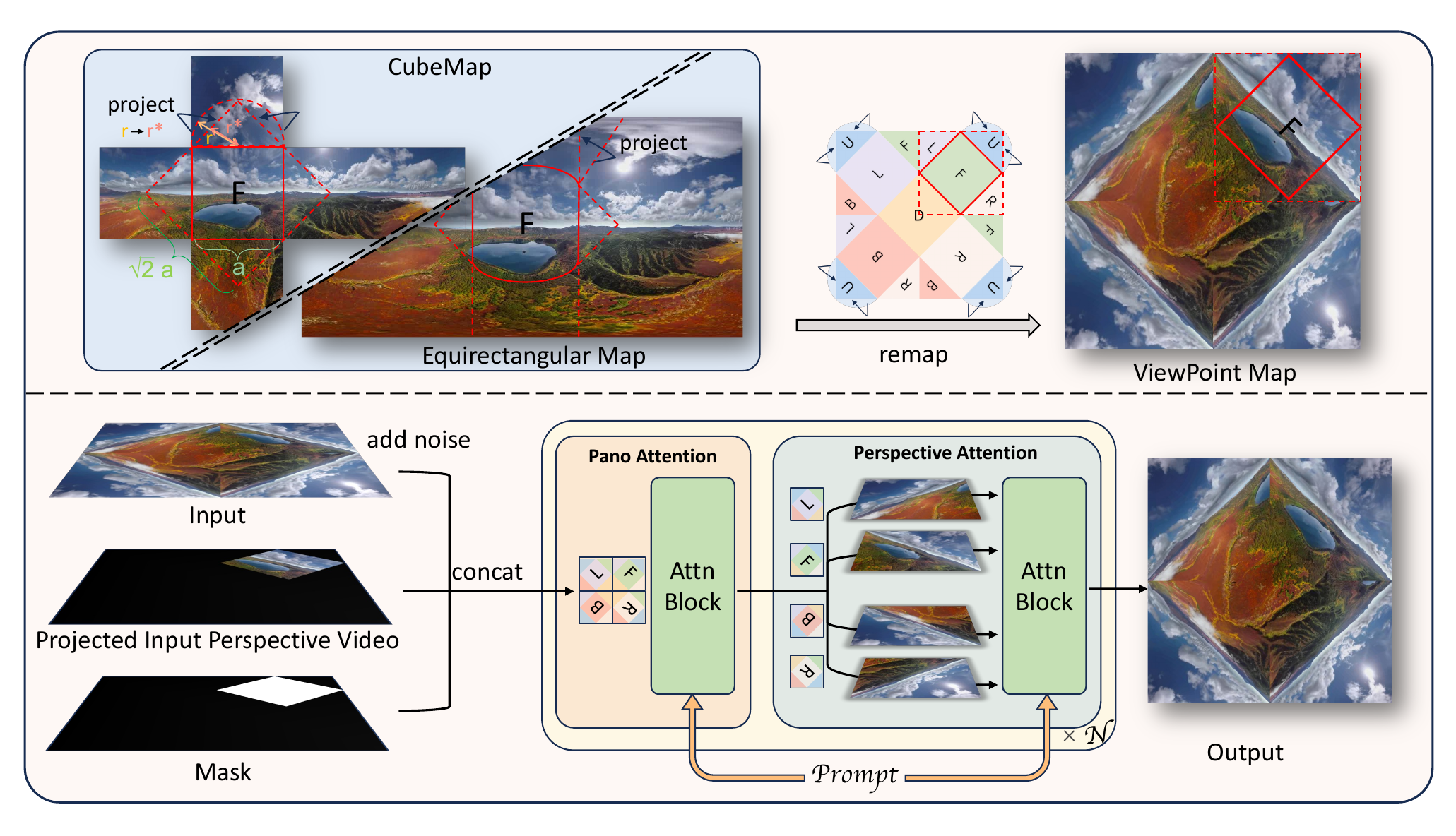}
    \vspace{-5.5mm}
  \caption{\textbf{Method overview.} The first row illustrates how to construct a Viewpoint map from a CubeMap or an Equirectangular Map. We begin by constructing subregions, using $\mathcal{F}$ as an example to create a pseudo-perspective region centered around it. Subsequently, we combine the four subregions to form a ViewPoint map. The second row shows the pipeline design: we first concatenate the noisy Viewpoint map, input video, and relevant mask along the channels dimension, and then employ a Pano-Perspective attention mechanism to learn how to maintain global spatial consistency while modeling fine-grained visual information.}
  \label{fig:pipe}
  % \vspace{-0.5cm}
\end{figure}

% To overcome the shortcomings mentioned above, we propose a new paradigm named ViewPoint,30
% which combines the advantages of ERP and CP. Specifically, we first project the panoramic views to31
% six cube faces with a FoV of 90◦ for each face, following the Cubemap format. This representation32
% ensures that each face is naturally suited for pre-trained diffusion schemes. Nevertheless, CP struggles33
% with spatial continuity due to the ordering of each cube face, as an unfolded cardboard box has at34
% most one face that is surrounded by four adjacent faces, thereby limiting the neural network’s ability35
% to model spatial consistency. To mitigate this limitation, we propose a pseudo-perspective view by36
% selecting one face as the central face (assuming the side length of r), then diagonally dividing each of 
% the four adjacent faces into four parts, as shown in Figure 1, and combining them into a new square 
% with the side length of √2r. This design makes the central face more spatially continuous with the 
% four adjacent faces. We then concatenate the four pseudo-perspective views centered on the left, front, 
% right, and back respectively, in raster order. The result is not only spatially consistent within each 
% pseudo-perspective region but also connects the entire cube space along the diagonals.

\subsection{ViewPoint Map}
%为了减少几何扭曲同时维持空间连续性，我们提出了...
To maintain spatial continuity while reducing spatial distortion, we propose ViewPoint Map, a novel panorama representation that combines the advantages of ERP and CP. 
We first sample six cube faces from either an equirectangular map or a cubemap, namely $\mathcal{F,R,B,L,U,}$ and $\mathcal{D}$, which represent the front, right, back, left, up, and down faces, respectively.
The four faces in the horizontal view, \ie, $\mathcal{F,R,B,L}$, typically contain most of the visual content that people focus on.
Therefore, we construct pseudo-perspective subregions centered around these four faces. 
As shown in the first row of ~\cref{fig:pipe}, we select one face as the central face (assuming the side length of $a$), then diagonally dividing each of the four adjacent faces into four parts, and finally, the central face and its four adjacent parts are concatenated to form a square region with a side length of $\sqrt[]2a$. 
This design makes the central face more spatially continuous with the four adjacent faces.
For example, if $\mathcal{F}$ is the central face, then its four adjacent faces are $\mathcal{L,R,U}$ and $\mathcal{D}$, located to the left, right, above, and below $\mathcal{L}$, respectively.
After obtaining the four subregions, we perform rotation and concatenation on them to further ensure that the center of the ViewPoint Map—the $\mathcal{D}$ face—is also continuous.
To address the splitting of $\mathcal{U}$, we project the semicircular region on the $\mathcal{U}$ face adjacent to the central face into an equilateral right triangle.
Formulally, for any point $P (r, \theta)$ on the semicircular region, where $r \in [0, a]$ and $\theta \in [0, \pi]$, with $a$ being the diameter of the semicircle. The scale ratio is defined as 
\begin{equation}  
d(\theta) = \frac{a}{sin\theta+\lvert cos\theta \rvert},
\end{equation}
then the scaled radial coordinate $r^*$ is computed by
\begin{equation}  
r^* = r \cdot \frac{d(\theta)}{a}.
\end{equation}
This design allows the $\mathcal{U}$ faces of the four subregions to have a small overlapping portion, thereby ensuring internal consistency of the entire $\mathcal{U}$ face.

\subsection{Pano-Perspective Attention}
ViewPoint Map offers global panorama information and effectively utilizes the inherent in-context generation capabilities~\cite{lhhuang2024iclora} of diffusion models. 
To further accelerate convergence and exploit the priors of the base model, we propose a Pano-Perspective Attention mechanism. 
Specifically, Pano-Attention is responsible for contextual learning to maintain spatial consistency, while Perspective-Attention focuses on generating fine-grained visual information for each subregion.
As shown in the second row of~\cref{fig:pipe}, we first concatenate the noisy Viewpoint Map, the projected input perspective video, and the associated mask along the channel dimension in the latent space.
The shape of the concatenated features is $(batch\_size, channels, frames, height, width)$. 
After passing through the Pano Attention block, we reshape it to $(4 \times batch\_size, channels, frames, height/2, width/2)$ and feed it into the Perspective Attention for per-regional modeling.
Each attention block consists of a self-attention layer, a cross-attention layer, and an FFN, the input prompt is integrated through the cross-attention mechanism.

\subsection{Overlapping Fusion}
\begin{figure}[!h]
  \centering
    % \vspace{-1.0cm}
  \includegraphics[width=0.8\linewidth]{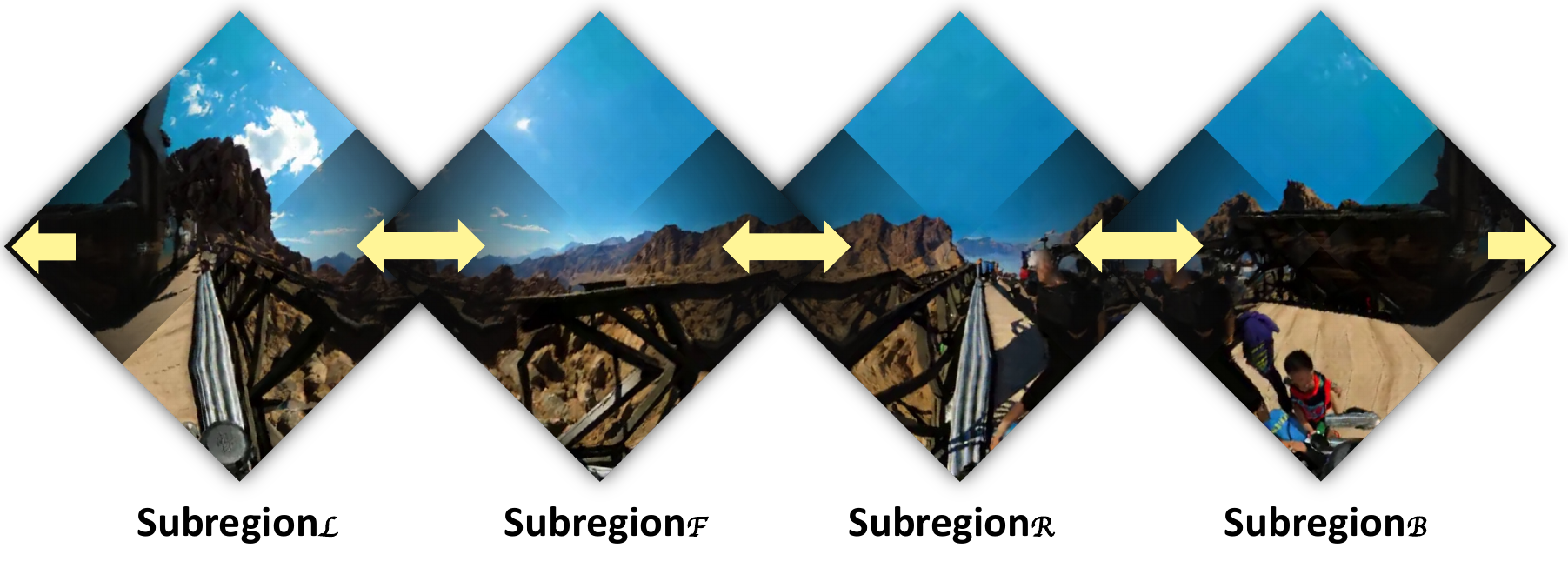}
  \caption{\textbf{Overlapping fusion.} The four subregions partially overlap with each other, thus we propose a gradient fusion mechanism to interpolate the overlapping areas, thereby enhancing spatial consistency.}
  \label{fig:fusion}
  %\vspace{-0.5cm}
\end{figure}

Each subregion in a ViewPoint map has a partial spatial overlap with the two adjacent subregions. 
To achieve smoother transitions between subregions, we propose a gradient fusion mechanism.
As shown in~\cref{fig:fusion}, the overlapping parts between two subregions form a regular rhombus, therefore we construct an overlapping fusion weight $W\in \mathbb{R}^{n\times n}$ where the values decay from the area near the central face to the edges.

For each subregion $S_d\in \mathbb{R}^{r\times r}$, $r=2n$, $d \in [\mathcal{L,F,R,B}]$, the overlapping fusion process can be described by the following formulas:
\begin{equation}  
W_{i,j} = \frac{i+j-2}{2(n-1)}, \quad i, j = 1, 2, \dots, n
\end{equation}
where $i, j$ denote the row and column indices, respectively. Next, we define rotation. For a given matrix $A$, we have
\begin{equation}  
R_{90}(A)= A^T \cdot J, %顺时针 
\end{equation}
\begin{equation}  
R_{-90}(A)=  J \cdot A^T ,  %逆时针
\end{equation}
\begin{equation}  
R_{180}(A)=  J \cdot A \cdot J ,  %180
\end{equation}
where $J$ is the exchange matrix. Finally, for each subregion, we have
% L
\begin{equation}  
S_{L_{i,j}}=\begin{cases}
S_{L_{i,j}} \cdot R_{90}(W) + R_{-90}(S_F)_{i+n,j-n} \cdot R_{-90}(W) \text{ , } (i,j)\in [1,n]\times [n+1,r] \\
S_{L_{i,j}} \cdot R_{-90}(W) + R_{90}(S_B)_{i-n,j+n} \cdot Rot_{90}(W) \text{ , } (i,j)\in [n+1,r]\times [1,n]\\
\end{cases}
\end{equation}
% R 
\begin{equation}  
S_{R_{i,j}}=\begin{cases}
S_{R_{i,j}} \cdot R_{90}(W) + R_{90}(S_F)_{i+n,j-n} \cdot R_{-90}(W) \text{ , } (i,j)\in [1,n]\times [n+1,r] \\
S_{R_{i,j}} \cdot R_{-90}(W) + R_{-90}(S_B)_{i-n,j+n} \cdot R_{90}(W) \text{ , } (i,j)\in [n+1,r]\times [1,n]\\
\end{cases}
\end{equation}
% F 
\begin{equation}  
S_{F_{i,j}}=\begin{cases}
S_{F_{i,j}} \cdot W + R_{90}(S_L)_{i+n,j+n} \cdot R_{180}(W) \text{ , } (i,j)\in [1,n]\times [1,n] \\
S_{F_{i,j}} \cdot R_{180}(W) + R_{-90}(S_R)_{i-n,j-n} \cdot W \text{ , } (i,j)\in [n+1,r]\times [n+1,r]\\
\end{cases}
\end{equation}
% B
\begin{equation}  
S_{B_{i,j}}=\begin{cases}
S_{B_{i,j}} \cdot W + R_{-90}(S_L)_{i+n,j+n} \cdot R_{180}(W) \text{ , } (i,j)\in [1,n]\times [1,n] \\
S_{B_{i,j}} \cdot R_{180}(W) + R_{90}(S_R)_{i-n,j-n} \cdot W \text{ , } (i,j)\in [n+1,r]\times [n+1,r]\\
\end{cases}
\end{equation}
Note that we apply overlapping fusion to each subregion simultaneously, rather than in the order specified in the above formulas.

For the $\mathcal{U}$ face, we first project the triangles back into semicircles and transform the rhombus into a petal shape formed by the overlap of two circles. 
Finally, we use a similar fusion mechanism to fuse the overlapping of the four semicircles.

%% file: secs/4_exp.tex
\section{Experiments}
% \subsection{Reconstruction Comparison}
\subsection{Implementation Details}
\noindent\textbf{Datasets \& Preprocess.} Our model is trained on 4 panorama datasets, including one image dataset, Flickr360~\cite{cao2023ntire}, and three video datasets, WEB360~\cite{wang2024360dvd}, ODV360~\cite{cao2023ntire}, and 360+x~\cite{chen2024x360}.
Among them, only WEB360~\cite{wang2024360dvd} comes with captions; therefore, we use Qwen-VL~\cite{Qwen-VL} to annotate the remaining three datasets, generating corresponding descriptive captions.
All datasets are resized to a resolution of $512 \times 1024$, and during training, video data is divided into clips of 49 frames each.

\noindent\textbf{Training.} We first inflate the patch embedding layer of the powerful video generation model, Wan2.1~\cite{wan2025}, from 16 to 33 channels to accommodate the input data, and then fine-tune the entire model.
The training process is executed on 8 $\times$ NVIDIA A100 GPUs, using a batch size of 1 and a learning rate of $1e-4$. 
We employ a joint image-video training strategy, treating images as videos with only one frame.

%\noindent\textbf{Inference.}% Optional 
\begin{figure}[t]
  \centering
   \vspace{-0.5cm}
  \includegraphics[width=1\linewidth]{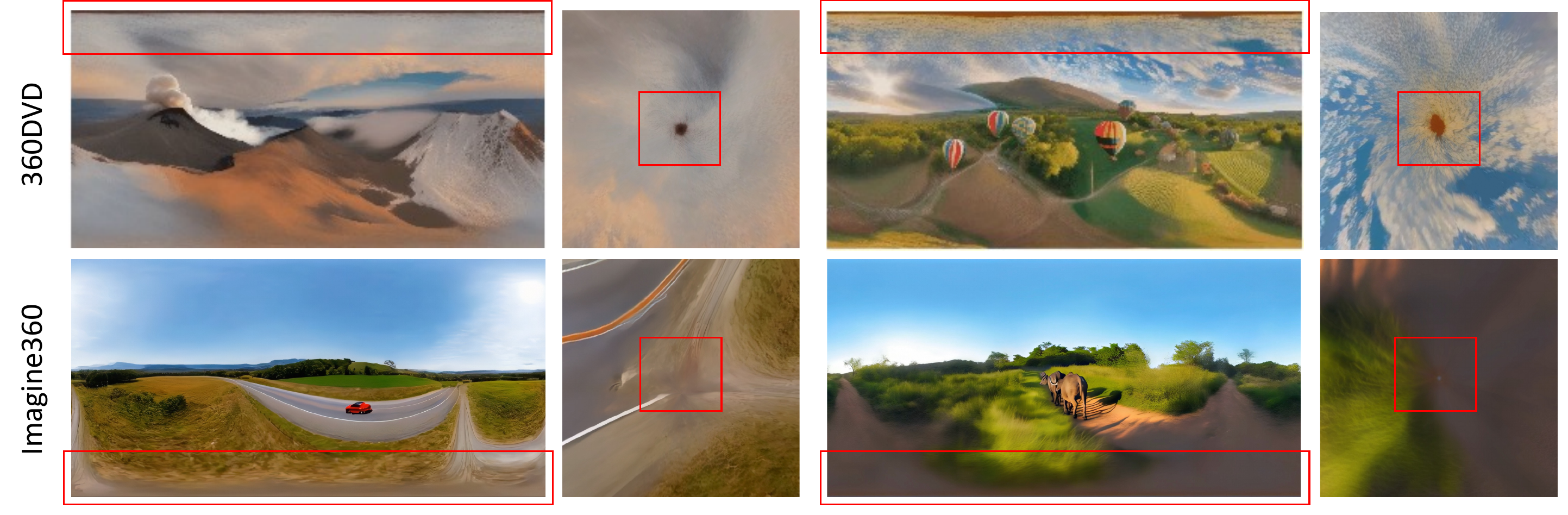}
  \caption{\textbf{Results from 360DVD~\cite{wang2024360dvd} and Imagine360~\cite{tan2024imagine360}.} Despite numerous efforts to mitigate geometric distortion at the poles, both methods still struggle with generating realistic top and bottom views. The distortion at the poles is even more pronounced in video scenes, severely affecting realism and immersion.}
  \label{fig:exp1}
  \vspace{-0.5cm}
\end{figure}
\subsection{Qualitative Comparison}
We compare our approach with three previous methods, including one perspective video outpainting method, Follow-Your-Canvas~\cite{chen2024follow}, and two 360-degree video generation methods, 360DVD~\cite{wang2024360dvd} and Imagine360~\cite{tan2024imagine360}. 
Adopting equirectangular representation, both 360DVD~\cite{wang2024360dvd} and Imagine360~\cite{tan2024imagine360} exhibit severe distortion at the poles.
As shown in~\cref{fig:exp1}, the sky appears to have ``black holes", while the ground shows ``swirls".
We argue that these artifacts are caused by the model's difficulty in handling the spatial-temporal distortion introduced by ERP, as a slight movement can result in significant disturbances at the poles.

\begin{figure}[t]
  \centering
   \vspace{-0.5cm}
  \includegraphics[width=1\linewidth]{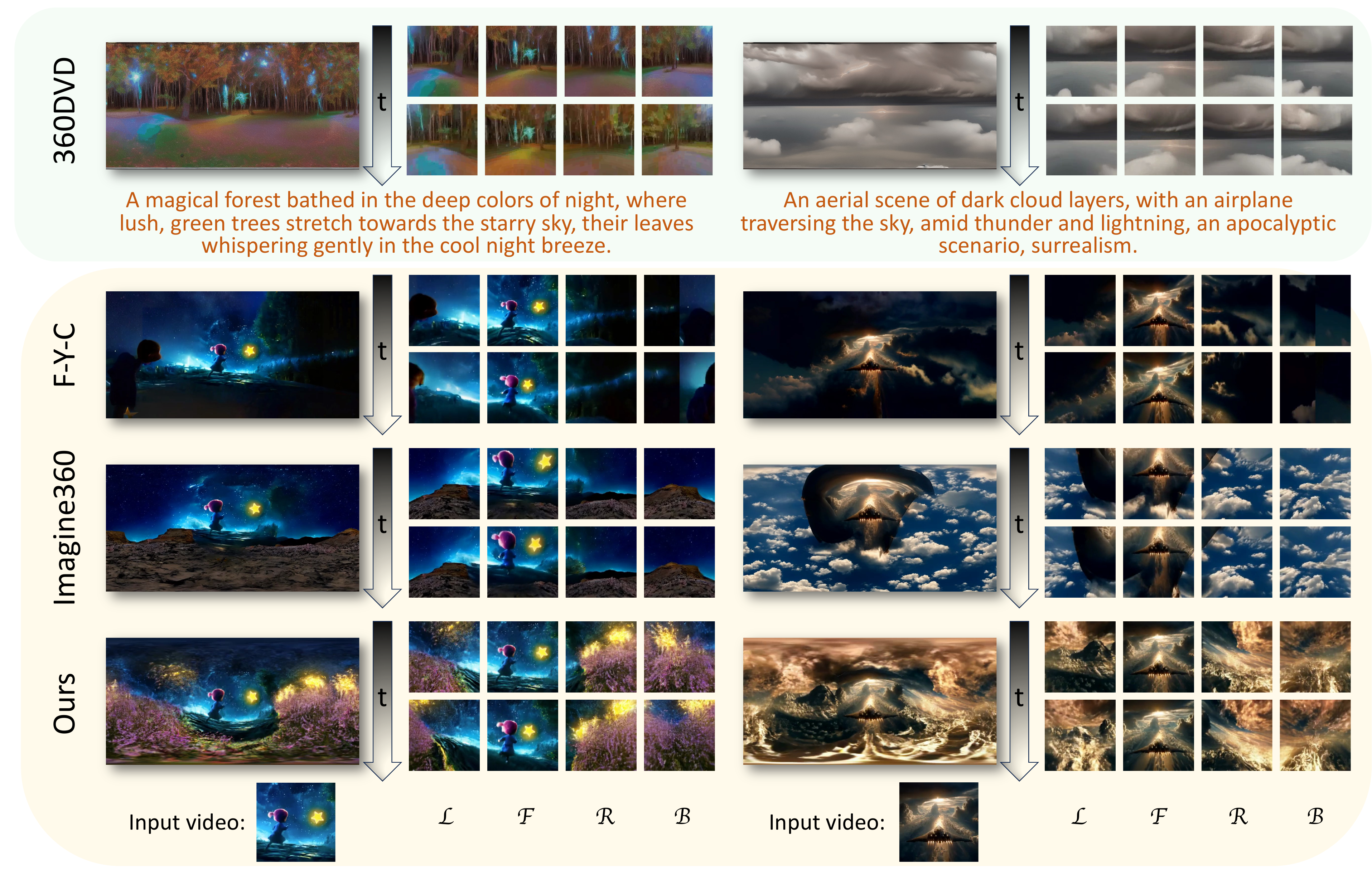}
  \caption{\textbf{Qualitative comparison of generated videos.} 360DVD is a text-driven approach and exhibits terrible image quality while Follow-Your-Canvas fails to generate panoramic videos with a reasonable spatial layout. Imagine360 suffers from spatial-temporal discontinuity. Our approach, on the other hand, can generate high-quality panoramic videos aligned with the given input. Due to page limits, we highly recommend watching the dynamic videos available at \href{https://becauseimbatman0.github.io/ViewPoint}{\textcolor{red}{ViewPoint}}.}
  \label{fig:compare}
  %\vspace{-0.5cm}
\end{figure}

The visualization of the comparison results is shown~\cref{fig:compare}, where we present the ERP format and perspective views in four horizontal directions, with arrows indicating the flow of time.
As shown in~\cref{fig:compare}, 360DVD~\cite{wang2024360dvd} only recognizes textual input. The generated frames exhibit terrible image quality and possess little motion.
Follow-Your-Canvas(F-Y-C)~\cite{chen2024follow} generates satisfactory perspective videos; however, its results suffer from severe spatial inconsistency, especially on the $\mathcal{B}$ face, exhibiting a noticeable sense of disconnection.
Although Imagine360~\cite{tan2024imagine360} synthesizes reasonable panoramic videos, it lacks spatial-temporal continuity.
As in the first example, the little girl is running, but the ground does not move accordingly and shows a noticeable texture difference from the input video.
In the second example, the aircraft in the input video is in a dark and tense atmosphere, while the generated panoramic video shows blue skies and white clouds, creating a peaceful scene. 
Additionally, the aircraft does not integrate well with the surrounding environment, resulting in a discordant effect.
In contrast, our approach demonstrates excellent spatial-temporal consistency and supports both textual prompts and input videos, thus enhancing the flexibility of generation. 
Furthermore, ViewPoint exhibits high motion dynamics, as each perspective view responds to the motion trends of the input video correctly.
As in the first case, the input video depicts a little girl running forward, with the prompt describing a magical forest scene at night. 
Our generated video aligns with both the input video and the prompt, showcasing a strong capability for condition awareness and integration.
Similarly, in the second example, the generated panoramic video also integrates with the input video and exhibits high-quality dynamics.

\subsection{Quantitative Comparison}
In this section, we provide a quantitative comparison of our approach with previous methods. 
VBench~\cite{huang2023vbench} is a comprehensive benchmark suite for video generative models which scores a video in 16 dimensions.
We evaluate our approach and previous methods on the ODV360~\cite{cao2023ntire} dataset across five dimensions: ``subject consistency", ``imaging quality", ``motion smoothness" and ``dynamic degree".
Specifically, we first project the generated videos into six perspective views with a FoV of 90$^\circ$, similar to $\mathcal{F,R,B,L,U,}$ and $\mathcal{D}$.
Then, we apply VBench~\cite{huang2023vbench} evaluation in a perspective manner for all projected videos.

The results of the quantitative evaluation are shown in~\cref{tab:vbench}, where our approach achieves the best scores across four metrics.
Although 360DVD~\cite{wang2024360dvd} performs the second best in ``subject consistency", it scores the lowest in ``dynamic degree", indicating that the videos generated by 360DVD have very little motion. 
The nearly static videos result in 360DVD achieving a high score in ``motion smoothness"; however, such videos fail to meet the demands of generating high-quality panoramic videos. 
Follow-Your-Canvas~\cite{chen2024follow}, on the other hand, is capable of generating high-quality perspective results, but the projected videos exhibit severe spatial distortion, resulting in poor performance in evaluations.
In contrast, our approach demonstrates state-of-the-art performance in all four metrics, proving that our method can generate highly dynamic and temporally coherent panoramic videos.

% \begin{table}[h]
% \setlength\tabcolsep{2pt}%
% \renewcommand\arraystretch{1.2}

%   \centering
%   % \vspace{-2mm}
%     \caption{\textbf{Quantitative comparison on VBench.} Our method achieves the best performance in four metrics and the second place in one metric. Although 360DVD obtains the second highest ``subject consistency" score due to the nearly static nature of its generated videos, the results fail to meet the demands of generating high-quality panoramic videos. In contrast, our approach performs well in both motion consistency and dynamics.}
%     % \vspace{2mm}
    
%     \resizebox{1\linewidth}{!}{
%    \begin{tabular}{c|ccccc}
%     \toprule
%     Method  & subject consistency$\uparrow$ & imaging quality$\uparrow$ & motion smoothness$\uparrow$ & aesthetic quality$\uparrow$ & dynamic degree$\uparrow$ \\
%     \midrule
%     Follow-Your-Canvas~\cite{chen2024follow} &0.8284 &0.4464 &0.9655 &0.4118 &0.8500 \\
%     360DVD~\cite{wang2024360dvd}  &0.8633 &0.5394 &0.9703  &0.4651 &0.5083\\
%     Imagine360~\cite{tan2024imagine360}  &0.8547 &0.5859 &0.9720  &\textbf{0.5313} &0.8148 \\
    
%     \textbf{\method(Ours)}  &\textbf{0.8793} &\textbf{0.5927} &\textbf{0.9800} &0.5052 &\textbf{0.9083}\\
%     \bottomrule
%   \end{tabular}
% }
%   \label{tab:vbench}
% \end{table}

\begin{table}[h]
\setlength\tabcolsep{2pt}%
\renewcommand\arraystretch{1.2}

  \centering
  % \vspace{-2mm}
    \caption{\textbf{Quantitative comparison on VBench.} Our method achieves the best performance in all four metrics. Although 360DVD obtains the second highest ``subject consistency" score due to the nearly static nature of its generated videos, the results fail to meet the demands of generating high-quality panoramic videos. In contrast, our approach performs well in both motion consistency and dynamics.}
    % \vspace{2mm}
    
    \resizebox{0.8\linewidth}{!}{
   \begin{tabular}{c|cccc}
    \toprule
    Method  & subject consistency$\uparrow$ & imaging quality$\uparrow$ & motion smoothness$\uparrow$ & dynamic degree$\uparrow$ \\
    \midrule
    Follow-Your-Canvas~\cite{chen2024follow} &0.8284 &0.4464 &0.9655  &0.8500 \\
    360DVD~\cite{wang2024360dvd}  &0.8633 &0.5394 &0.9703   &0.5083\\
    Imagine360~\cite{tan2024imagine360}  &0.8547 &0.5859 &0.9720  &0.8148 \\
    
    \textbf{\method(Ours)}  &\textbf{0.8793} &\textbf{0.5927} &\textbf{0.9800} &\textbf{0.9083}\\
    \bottomrule
  \end{tabular}
}
  \label{tab:vbench}
\end{table}

\subsection{Ablation Study}
To demonstrate the effectiveness of our method, we conduct ablation experiments on panorama representation formats and network designs.
For the ERP format, we directly fine-tune the base model to adapt to this panoramic representation. 
For the CP format, we train two models, CP-ICLoRA and CP-4DRoPE, separately to assess the impact of different cubeface-encodings.
Technically, CP-ICLoRA rearranges the six cube faces into a 2-row by 3-column rectangle in raster order, \ie $\mathcal{F,R,B}$ on the first row, and $\mathcal{L,U,D}$ on the second row, while CP-4DRoPE assigns an independent positional encoding to each cube face.
We also examine the design of Pano-Perspective attention by replacing all Perspective blocks with the intact Pano blocks.

\begin{figure}[t]
  \centering
    \vspace{-0.5cm}
  \includegraphics[width=1\linewidth]{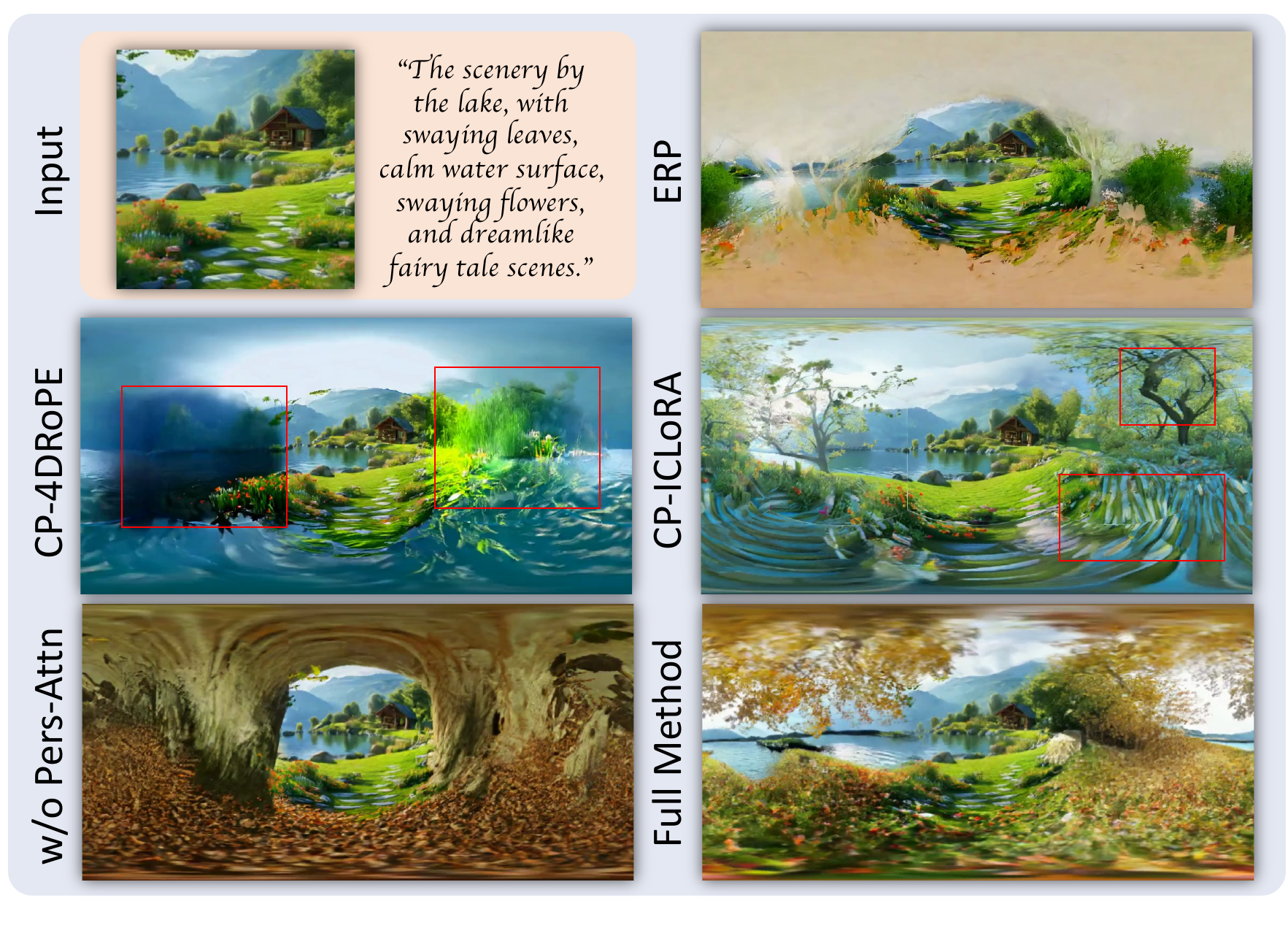}
   \vspace{-8.0mm}
  \caption{\textbf{Ablation on different designs.} ERP exhibits serious artifacts due to the natural gap in modality. Both Cube representation methods have spatial discontinuity issues. Without Perspective-Attention, it leads to misalignment with the input video. Our full method can generate reasonable and spatially consistent results.}
  \label{fig:ablation}
  \vspace{-5.5mm}
\end{figure}

As shown in~\cref{fig:ablation}, ERP exhibits significant artifacts due to the inherent modality gap with the pretrained model. 
Both cubemap representation methods suffer from spatial discontinuity, wherein CP-4DRoPE shows inconsistency in color tone, while CP-ICLoRA, although achieving better image quality, presents significant spatial fractures.
Our full method, in contrast, is capable of generating high-quality and coherent panoramic videos that are superior to other alternative designs in both spatial and temporal aspects.

The results of quantitative ablation are shown in~\cref{tab:ablation}, our full method outperforms the other four alternative designs across four metrics.
Since we conduct evaluations in a perspective manner, CP-ICLoRA performs best in ``imaging quality''. However, as shown in~\cref{fig:ablation}, though each perspective video shows good visual quality, there is still a discontinuity issue at the boundaries of the cube faces. 

% \begin{table}[h]
% \setlength\tabcolsep{2pt}%
% \renewcommand\arraystretch{1.2}

%   \centering
%   % \vspace{-2mm}
%     \caption{\textbf{Quantitative ablation.} Compared to the other four alternative designs, our approach achieves the highest scores across four metrics.}
%     % \vspace{2mm}
    
%     \resizebox{1\linewidth}{!}{
%    \begin{tabular}{c|ccccc}
%     \toprule
%        & subject consistency$\uparrow$ & imaging quality$\uparrow$ & motion smoothness$\uparrow$ & aesthetic quality$\uparrow$ & dynamic degree$\uparrow$ \\
%     \midrule
%     ERP &0.8702 &0.5396 &0.9781 &0.4913 &0.8840\\
%     CP-4DRoPE  &0.8612 &0.5104 &0.9633  &0.5012 &0.8175  \\
%     CP-ICLoRA  &0.8727 &\textbf{0.6018} &0.9786  &0.4930 &0.8946  \\
%     w/o Perspective-Attn  &0.8663 &0.5392 &0.9798  &0.4254 &0.8827 \\
%     \textbf{Full Method}  &\textbf{0.8793} &0.5927 &\textbf{0.9800} &\textbf{0.5052} &\textbf{0.9083} \\
%     \bottomrule
%   \end{tabular}
% }
%   \label{tab:ablation}
% \end{table}
\begin{table}[h]
\setlength\tabcolsep{2pt}%
\renewcommand\arraystretch{1.2}

  \centering
  % \vspace{-2mm}
    \caption{\textbf{Quantitative ablation.} Compared to the other four alternative designs, our approach achieves the highest scores across three metrics. Although CP-ICLoRA achieves the highest score in ``imaging quality", proving that each of the six faces has good quality independently, the panoramas composed of these six faces exhibit noticeable discontinuities, as indicated by the red boxes in~\cref{fig:ablation}. Therefore, it does not meet the demand for high-quality panoramic video generation.}
    % \vspace{2mm}
    
    \resizebox{0.8\linewidth}{!}{
   \begin{tabular}{c|ccccc}
    \toprule
       & subject consistency$\uparrow$ & imaging quality$\uparrow$ & motion smoothness$\uparrow$ & dynamic degree$\uparrow$ \\
    \midrule
    ERP &0.8702 &0.5396 &0.9781 &0.8840\\
    CP-4DRoPE  &0.8612 &0.5104 &0.9633  &0.8175  \\
    CP-ICLoRA  &0.8727 &\textbf{0.6018} &0.9786   &0.8946  \\
    w/o Perspective-Attn  &0.8663 &0.5392 &0.9798  &0.8827 \\
    \textbf{Full Method}  &\textbf{0.8793} &0.5927 &\textbf{0.9800}  &\textbf{0.9083} \\
    \bottomrule
  \end{tabular}
}
  \label{tab:ablation}
\end{table}
\subsection{User Study}
Despite achieving advanced scores on VBench~\cite{huang2023vbench}, we conduct user studies, introducing subjective user ratings to further validate the superiority of our approach.
We ask participants to vote on the tested videos based on four dimensions: ``Spatial Continuity", ``Temporal Quality", ``Aesthetic Preference" and ``Condition Alignment". ``Spatial continuity" represents the coherence of scenes and structures in panoramic space, while ``Temporal Quality" is used to evaluate motion effects; for example, being nearly motionless and flickering are considered poor performance. ``Aesthetic Preference" represents participants' subjective preferences, while ``Condition Alignment" refers to the degree to which the generated panoramic video aligns with the input conditions, such as text or video.
%
% \begin{wrapfigure}{r}{0.5\linewidth}
%     % \vspace{-2.5mm}
%     \includegraphics[width=0.9\linewidth]{figs/user.pdf}
%     \caption{\textbf{User studies.} We ask participants to vote on the videos generated by four methods based on four dimensions, and our approach receives the most votes.}
%      % \vspace{-15.0mm}
%     \label{fig:user}
% \end{wrapfigure}

Finally, we collect 50 valid questionnaires, each containing 14 sets of videos for comparison, and the results are shown in~\cref{fig:user}.
Overall, Imagine360~\cite{tan2024imagine360} is the second choice among participants, while 360DVD~\cite{wang2024360dvd} and Follow-Your-Canvas~\cite{chen2024follow}, receive the fewest votes.
Our approach receives the highest number of votes across all four dimensions, surpassing 360DVD~\cite{wang2024360dvd}, Follow-Your-Canvas~\cite{chen2024follow}, and Imagine360~\cite{tan2024imagine360}, demonstrating our method's ability to generate satisfactory panoramic results.
%

% \begin{figure}[h]
%   \centering
%     % \vspace{-1.0cm}
%   \includegraphics[width=0.6\linewidth]{figs/user.pdf}
%   \caption{\textbf{User studies.} We ask participants to vote on the videos generated by four methods based on four dimensions, and our approach receives the most votes.}
%   \label{fig:user}
%   %\vspace{-0.5cm}
% \end{figure}

%% file: secs/5_conclusion.tex
\section{Conclusion}
\begin{wrapfigure}{r}{0.5\linewidth}
    \vspace{-2.5mm}
    \includegraphics[width=0.9\linewidth]{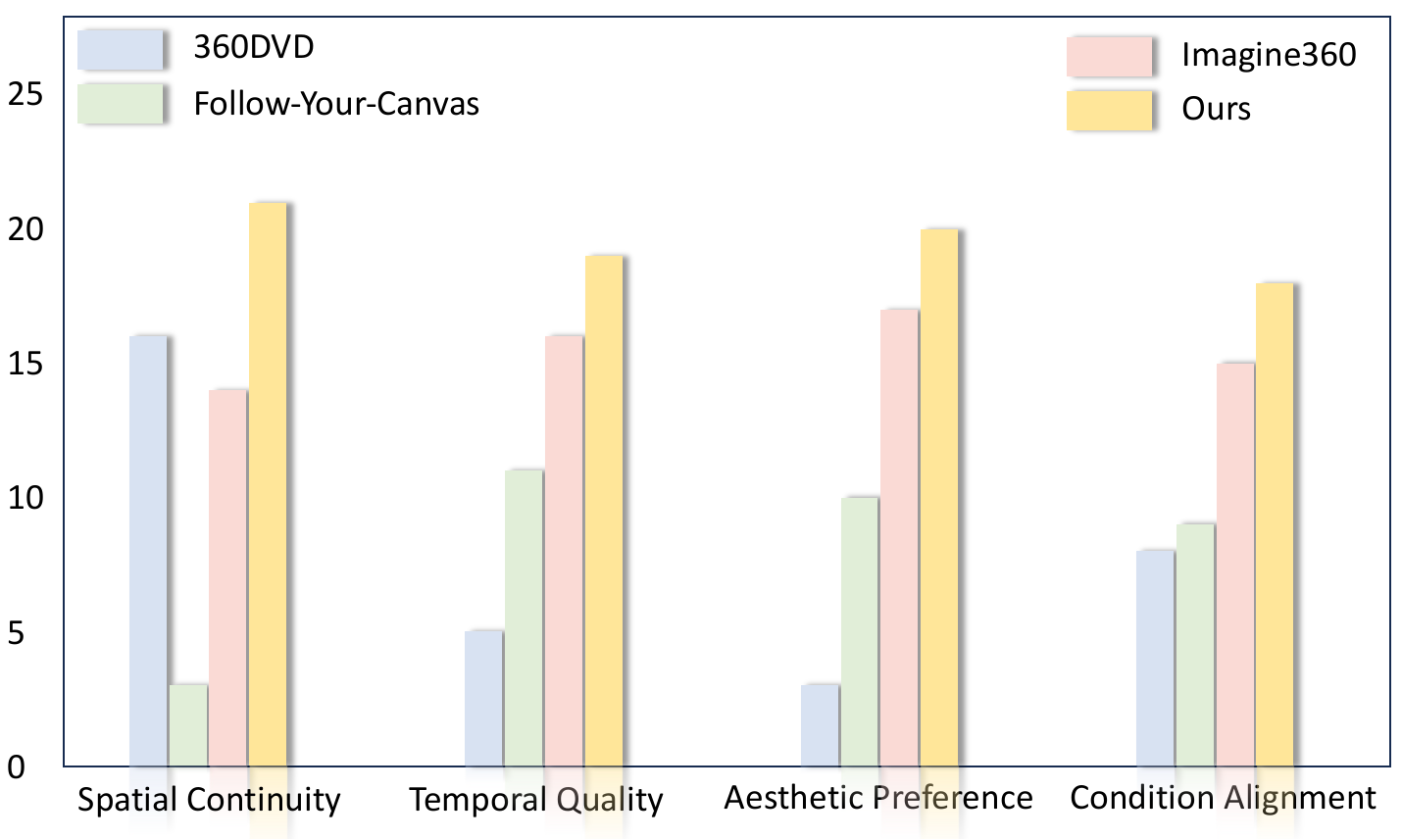}
    \caption{\textbf{User studies.} We ask participants to vote on the videos generated by four methods based on four dimensions, and our approach receives the most votes.}
     % \vspace{-10.0mm}
    \label{fig:user}
\end{wrapfigure}
In this work, we present ViewPoint, a novel framework for representing and generating panoramic videos leveraging modern generative models. 
Specifically, we design a novel representation distinct from traditional panoramic image and video representations, which has the advantage of good spatial continuity and temporal consistency.
Through our proposed Pano-Perspective attention mechanism, the pretrained model perceives the global spatial structure information of panoramic videos while modeling fine-grained visual features, effectively improving the quality of the generated videos.
To further enhance spatial consistency, we propose an overlapping gradient fusion mechanism that fully utilizes the spatial continuity of each subregion, thereby improving the spatial quality of panoramic videos.
Extensive qualitative and quantitative experiments, as well as user studies, demonstrate the effectiveness of our method, and we believe ViewPoint can provide valuable insights for the omnidirectional vision community.

%% file: secs/6_ref.tex
{
\small
\bibliographystyle{plainnat}
\bibliography{ref.bib}
}